\documentclass[a4paper,fleqn]{cas-dc}
\usepackage[numbers,sort&compress]{natbib}
\usepackage{mathtools}
\usepackage{placeins}
\usepackage{microtype}
\usepackage{etoolbox}
\usepackage{xurl}
\usepackage{flushend}
\allowdisplaybreaks[0]
\usepackage{xcolor}
\hypersetup{colorlinks=true,linkcolor=DarkSlateGrey,citecolor=DarkSlateGrey,urlcolor=DarkSlateGrey,
 pdftitle={CGDD-Net: Context-Guided Dynamic Detail Modeling for Retinal Vessel Segmentation},
 pdfauthor={Xincheng Li; Xiaoqi Sheng; Xinyu Zhang},
 pdfsubject={Retinal vessel segmentation with context-guided dynamic detail modeling}}

\newcommand{\CGDDTableStyle}{\fontsize{7.2}{8.7}\selectfont\renewcommand{\arraystretch}{1.12}}
\newcommand{\CGDDTableHeadStrut}{\rule{0pt}{11pt}}

\newcommand{\CGDDTableSubHeadStrut}{\rule{0pt}{10pt}}
\newcommand{\CBR}{\operatorname{CBR}}

\newcommand{\Proj}{\operatorname{Proj}}
\newcommand{\Dec}{\operatorname{Dec}}
\newcommand{\CCA}{\operatorname{CCA}}

\begin{document}
\let\WriteBookmarks\relax
\renewcommand{\figurename}{Fig.}
\shorttitle{CGDD-Net: Context-Guided Dynamic Detail Modeling for Retinal Vessel Segmentation}
\shortauthors{X. Li et al.}
\title[mode=title]{CGDD-Net: Context-Guided Dynamic Detail Modeling for Retinal Vessel Segmentation}
\author[1]{Xincheng Li}[orcid=0009-0007-2730-2737]
\ead{xli798@aucklanduni.ac.nz}
\credit{Conceptualization, Methodology, Software, Validation, Formal analysis, Investigation, Data curation, Visualization, Writing -- original draft, Writing -- review \& editing}

\author[2]{Xiaoqi Sheng}
\cormark[1]
\ead{xqsheng@scut.edu.cn}
\credit{Supervision, Methodology, Validation, Writing -- review \& editing}

\author[1]{Xinyu Zhang}
\cormark[1]
\ead{xinyu.zhang@auckland.ac.nz}
\credit{Validation, Formal analysis, Investigation, Writing -- review \& editing}
\affiliation[1]{organization={School of Computer Science},addressline={The University of Auckland},city={Auckland},postcode={1010},country={New Zealand}}

\affiliation[2]{organization={School of Future Technology},addressline={South China University of Technology},city={Guangzhou},postcode={510641},state={Guangdong},country={China}}

\cortext[1]{Corresponding authors: Xiaoqi Sheng and Xinyu Zhang.}

\begin{abstract}
Accurate retinal vessel segmentation requires features that capture vascular geometry while preserving information for fine-scale reconstruction. We propose CGDD-Net, a context-guided dynamic detail modeling network that connects adaptive feature extraction to a shared decoder pathway. Context-Guided Scale-Adaptive Deformable Encoding (CSDE) combines fixed-grid convolution with deformable local attention to capture vascular patterns at multiple spatial extents. Spatially Adaptive Multi-Kernel Gating (SAMG) selects receptive-field responses at each location. Dynamic Cross-Scale Detail Fusion (DCDF) aligns the gated intermediate features and compresses them into an eight-channel representation, which is reused at three decoder resolutions together with selected encoder skips. This design consolidates intermediate information before decoding instead of transferring each middle-stage feature through a separate direct skip. On DRIVE, CHASE\_DB1, STARE, and HRF, CGDD-Net achieves AUC values of 0.9824, 0.9938, 0.9895, and 0.9874, with F1 scores of 0.8323, 0.8102, 0.8510, and 0.8157, respectively. The complete model contains 1.96 million trainable parameters. In cumulative ablations, the full model improves F1 over the internal baseline by 2.48, 0.61, and 3.49 percentage points on DRIVE, CHASE\_DB1, and STARE. Twelve directed cross-dataset experiments further characterize transfer without target-domain adaptation. The results support shared intermediate detail delivery as an effective, parameter-compact architecture for retinal vessel segmentation. Code is available at \url{https://github.com/lixincheng-xcl/CGDD-Net}.
\end{abstract}

\begin{keywords}
Retinal vessel segmentation \sep Fundus imaging \sep Deformable attention \sep Multi-scale fusion \sep Detail-guided decoding
\end{keywords}
\maketitle

\hypersetup{pdfauthor={Xincheng Li; Xiaoqi Sheng; Xinyu Zhang},
 pdfsubject={Retinal vessel segmentation with context-guided dynamic detail modeling},
 pdfkeywords={Retinal vessel segmentation, fundus imaging, deformable attention, multi-scale fusion, detail-guided decoding}}

\section{Introduction}\label{sec:introduction}
Retinal vessel maps support the measurement of caliber, tortuosity, and branching patterns in fundus images \cite{Qin2024Review,Staal2004DRIVE,Fraz2012CHASE}. Automatic segmentation reduces labor-intensive manual delineation and provides a consistent anatomical representation for subsequent vascular analysis. Its value depends on more than identifying the prominent vessel trunks. Small branches contribute to the spatial organization of the vascular tree, while accurate boundaries determine the apparent width of each segment. A missing branch and an enlarged vessel boundary introduce different errors into this representation, even when the main vessels remain recognizable. Accurate segmentation must therefore retain fine spatial information while using context to distinguish weak vessel responses from surrounding retinal structures.

\begin{figure}[pos=t]
\centering\includegraphics[width=0.88\columnwidth]{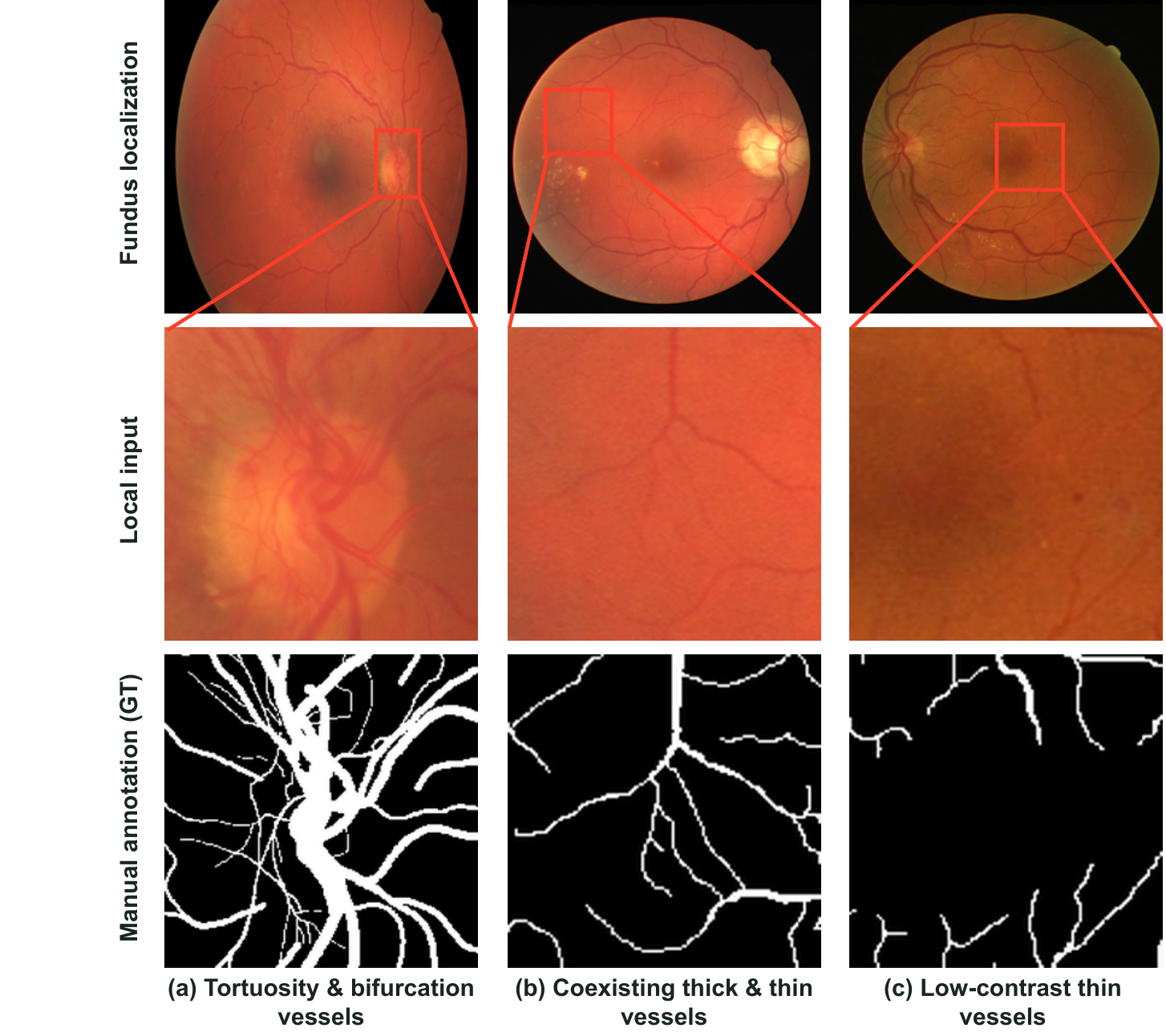}
\caption{Representative image structures motivating CGDD-Net: (a) tortuous vessels and bifurcations, (b) coexisting thick and thin vessels, and (c) low-contrast thin vessels. From top to bottom, each column shows the fundus image, the region indicated by the red box, and its manual vessel annotation (GT).}
\label{fig:motivation}
\end{figure}

The examples in Fig.~\ref{fig:motivation} illustrate how these requirements arise within individual images. Tortuous vessels change direction over short distances, and a bifurcation presents several orientations within the same neighborhood. Thick and thin vessels can also run close together, requiring different spatial extents to resolve their boundaries. For a faint branch, evidence along its course may be more informative than the intensity at one location. Simply enlarging the neighborhood does not address every case: the useful support depends on the geometry and contrast of the structure being reconstructed.

Encoder-decoder networks, including U-Net and its variants, combine spatial detail with contextual information \cite{Ronneberger2015UNet,Gu2019CENet,Zhuang2019LadderNet,Li2020IterNet,Mou2021CS2}. Deformable operations adapt sampling locations to the input, while attention weights the sampled features \cite{Dai2017Deformable,Jin2019DUNet,Li2020MAUNet,Xia2022DAT}. Multiple receptive fields help represent vessels of varying widths \cite{Wu2021SCSNet,Lin2021MPSNet,Ye2022MFINet,Zhu2025AMFEF,Cao2025MSVENet}. These mechanisms determine where to collect evidence and over what spatial extent. Convolution on the fixed grid retains local cues alongside these adaptive responses, supplying the decoder with complementary features for structures within each image.

\suppressfloats[t]

A separate architectural question is how this information reaches reconstruction. A direct skip transfers one encoder feature map to its matching decoder stage for fusion with an upsampled representation. The connection pattern thus shapes what is available to each stage. Middle encoder features lie between shallow representations for precise localization and deeper representations of broader context. Consolidating selected responses from these intermediate levels brings their scales together before decoding. This changes their transfer from a set of isolated skip inputs to multi-scale information that successive reconstruction stages can use.

We build a compact representation once and reuse it during reconstruction. This pathway carries selected middle-stage information alongside the shallow and deep direct connections. Separate decoder transformations accommodate the transition from coarse reconstruction to finer spatial detail. Our focus is this shared intermediate representation. It provides a common source of multi-scale features while leaving each decoder stage free to learn its own resolution-specific projection.

The proposed network forms one eight-channel detail representation from three middle encoder stages. Context-Guided Scale-Adaptive Deformable Encoding (CSDE) supplies local and geometry-adaptive features, and Spatially Adaptive Multi-Kernel Gating (SAMG) weights their receptive-field responses at each location. Dynamic Cross-Scale Detail Fusion (DCDF) combines the gated features on a common grid. The resulting representation guides three decoder resolutions through independent projections. Direct shallow and deep skips are retained for full-resolution localization and semantic context, respectively. Thus, shared consolidation does not replace every encoder-decoder connection; it reorganizes the delivery of the middle-stage information. Here, detail denotes the intermediate feature representation used for vascular reconstruction, rather than a separately supervised anatomical output.

The contributions of this work are:
\begin{enumerate}
\item We propose a shared detail-guided architecture that consolidates three intermediate encoder features into one fixed-width representation and reuses it at three decoder stages. Selective direct skips connect this pathway to full-resolution localization and deeper context.
\item We integrate geometry-adaptive encoding with spatially varying multi-kernel selection before cross-scale fusion. This connects the selection of sampling locations and receptive fields to the information delivered during reconstruction.
\item We evaluate CGDD-Net on four retinal vessel datasets, using cumulative ablations, qualitative comparisons, hyperparameter analysis, and twelve directed cross-dataset tests. The results characterize the segmentation performance and model-size profile of the complete 1.96M-parameter network.
\end{enumerate}

Section~\ref{sec:related} reviews related work, and Section~\ref{sec:methods} defines CGDD-Net. The experiments and their interpretation follow in Sections~\ref{sec:experiments} and \ref{sec:discussion}; Section~\ref{sec:conclusion} concludes the paper.

\section{Related work}\label{sec:related}
\subsection{Retinal vessel segmentation based on CNNs}
Early vessel segmentation relied on appearance filters and geometric descriptors \cite{Hoover2000STARE,Staal2004DRIVE,Fraz2012CHASE}. Deep networks subsequently learned vascular representations directly from annotated images \cite{Liskowski2016DNN,Yan2018JointLoss}. U-Net established a widely used encoder-decoder structure with high-resolution skips \cite{Ronneberger2015UNet}. Later designs organized contextual extraction and reconstruction in different ways: LadderNet extends feature reuse through multiple paths, CE-Net enlarges contextual support, and IterNet repeatedly refines vascular predictions \cite{Zhuang2019LadderNet,Gu2019CENet,Li2020IterNet}. Both additional paths and repeated refinement make information acquired at different depths available to reconstruction.

Geometry-aware methods address the spatial support of feature extraction. DUNet introduces deformable convolution for retinal vessels, MAU-Net combines adaptive operations with attention, and CS$^{2}$-Net models spatial and channel relationships \cite{Jin2019DUNet,Li2020MAUNet,Mou2021CS2}. CSDE combines deformable local attention with fixed-grid convolution before the features enter the shared detail pathway.

Multi-scale retinal models address variation in vessel appearance and width. SCS-Net combines scale-aware aggregation with adaptive fusion between adjacent hierarchical features and multi-level semantic supervision \cite{Wu2021SCSNet}. MPS-Net exchanges information between resolutions, MFI-Net couples interaction with refinement, and DPF-Net progressively fuses dual-path features \cite{Lin2021MPSNet,Ye2022MFINet,Li2023DPFNet}. AMFEF-Net and MSVENet also develop multi-scale vascular representations \cite{Zhu2025AMFEF,Cao2025MSVENet}. Extracting multi-scale responses and delivering them to reconstruction are distinct operations. CGDD-Net connects them by selecting responses within each intermediate stage before cross-stage fusion.

Other designs emphasize dual-path localization, spatial attention, compact convolutional blocks, or threshold selection \cite{Guo2020DPN,Guo2020SAUNet,Han2022ConvUNeXt,Getahun2023FSNet,Zeng2025AttUKAN}. Compactness is a separate architectural consideration. M2U-Net uses lightweight encoder-decoder blocks, while the minimalistic models of Galdran et al. show that narrow networks can achieve strong retinal segmentation scores \cite{Laibacher2018M2UNet,Galdran2022Minimalistic}. We likewise report model size alongside accuracy.

\subsection{Feature fusion based on deep learning}
Feature fusion determines which representations reach a reconstruction stage and how they are combined. U-Net concatenates features at matching scales; UNet++ introduces nested, dense skip pathways to reduce the semantic gap between encoder and decoder features; and U-Net3+ aggregates full-scale information at multiple decoder stages \cite{Ronneberger2015UNet,Zhou2018UNetPlusPlus,Huang2020UNet3Plus}. The connections and their transformations jointly determine the information delivered. Attention U-Net and AG-Net select features during transfer \cite{Oktay2018AttentionUNet,Zhang2019AGNet}. Transformer-based models introduce broader contextual interactions into encoder-decoder processing \cite{Chen2024TransUNet,Cao2023SwinUNet}, extending the information that can be combined beyond a purely local convolutional neighborhood.

A shared fused feature offers another way to organize this transfer. Cross-scale aggregation is computed once, while the decoder retains separate transformations for its different output resolutions.

Input-dependent fusion can also operate over parallel receptive fields. Selective Kernel Networks use attention to combine responses from branches with different kernel sizes, whereas dynamic convolution aggregates multiple kernels according to the input \cite{Li2019SelectiveKernel,Chen2020DynamicConv}. SAMG performs selection in feature space rather than generating a new convolution kernel at each pixel. Its pooled context descriptor and spatial logits jointly determine branch weights, allowing locations within the same feature map to use different mixtures of receptive-field responses. The selected features retain their encoder resolutions until DCDF aligns them.

Retinal-specific methods such as MFI-Net, DPF-Net, and MDF-Net combine multi-scale processing with refinement or attention-guided fusion \cite{Ye2022MFINet,Li2023DPFNet,Ma2025MDFNet}. CGDD-Net focuses on the subsequent delivery of intermediate responses: features from $E_2$, $E_3$, and $E_4$ are fused once at a fixed width and reused through stage-specific projections. Reuse therefore concerns a computed feature map, not tied projection weights. Together with the retained shallow and deep direct skips, this arrangement provides a shared route for middle-stage information rather than recomputing an independent cross-scale fusion at every reconstruction stage.

Structural supervision is a separate design dimension. Segment-level objectives and clDice explicitly incorporate structural information into optimization \cite{Yan2018JointLoss,Shit2021clDice}. CGDD-Net uses FOV-masked binary cross-entropy, concentrating the proposed changes in feature extraction and transfer. Its cumulative architectural comparisons therefore retain a common segmentation objective.

\section{Method}\label{sec:methods}
\subsection{Overall architecture}
\begin{figure*}[pos=t]
\centering\includegraphics[width=\textwidth]{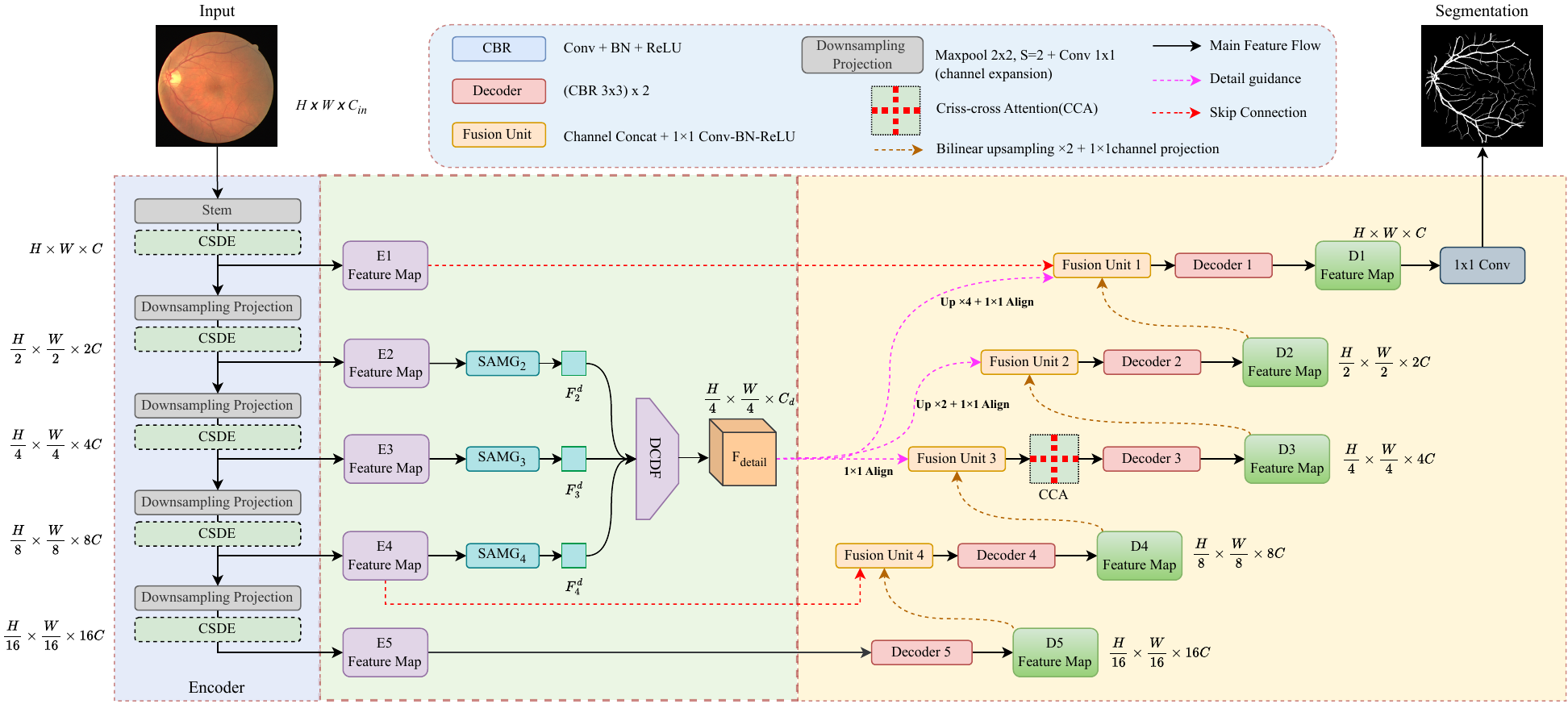}
\caption{CGDD-Net architecture with a shared intermediate detail pathway. CSDE generates encoder features $E_1$--$E_5$. SAMG selects multi-kernel responses from $E_2$, $E_3$, and $E_4$, which DCDF fuses into $F_{\mathrm{detail}}$ at quarter input resolution with $C_d=8$. Magenta dashed arrows route this feature to D3, D2, and D1 through independent projections; red dashed arrows denote the retained E4 and E1 skips. A single criss-cross attention (CCA) block follows D3 fusion. CBR denotes convolution, batch normalization, and ReLU. The final $1\times1$ convolution produces vessel logits.}
\label{fig:architecture}
\end{figure*}

CGDD-Net uses a five-stage encoder and decoder, as shown in Fig.~\ref{fig:architecture}. Let $E_i$ and $D_i$ denote features at resolution level $i$, ordered from finest to coarsest. For an input of size $H\times W$, the nominal spatial size at level $i$ is $H/2^{i-1}\times W/2^{i-1}$, and the channel widths are $8,16,32,64,128$. The shared detail representation has width $C_d=8$.

Tensor shapes are written in channel-first order, $C\times H\times W$, with the batch dimension omitted; the spatial-first labels in Fig.~\ref{fig:csde} describe the same tensors. Throughout this section, $\Vert$ denotes channel concatenation and $\CBR_k$ denotes a $k\times k$ convolution followed by batch normalization and ReLU.

A $3\times3$ CBR stem maps the grayscale input to eight channels. Each stage applies CSDE; adjacent stages are connected by $2\times2$ max-pooling and a $1\times1$ CBR channel projection. Encoding proceeds from $E_1$ to $E_5$, and decoding proceeds from $D_5$ to $D_1$.

The separate detail pathway processes $E_2$, $E_3$, and $E_4$ with SAMG, aligns the outputs to the $E_3$ grid, and fuses them with DCDF. The resulting feature is reused at D3, D2, and D1. Direct skips from $E_4$ and $E_1$ provide deeper encoder context and full-resolution localization. Intermediate features $E_2$ and $E_3$ reach the decoder only through the detail pathway.

Sharing refers to reuse of one computed feature map, not tied weights across decoder projections.

\subsection{Context-guided scale-adaptive deformable encoding}\label{sec:csde}
\begin{figure*}[pos=t]
\centering\includegraphics[width=\textwidth]{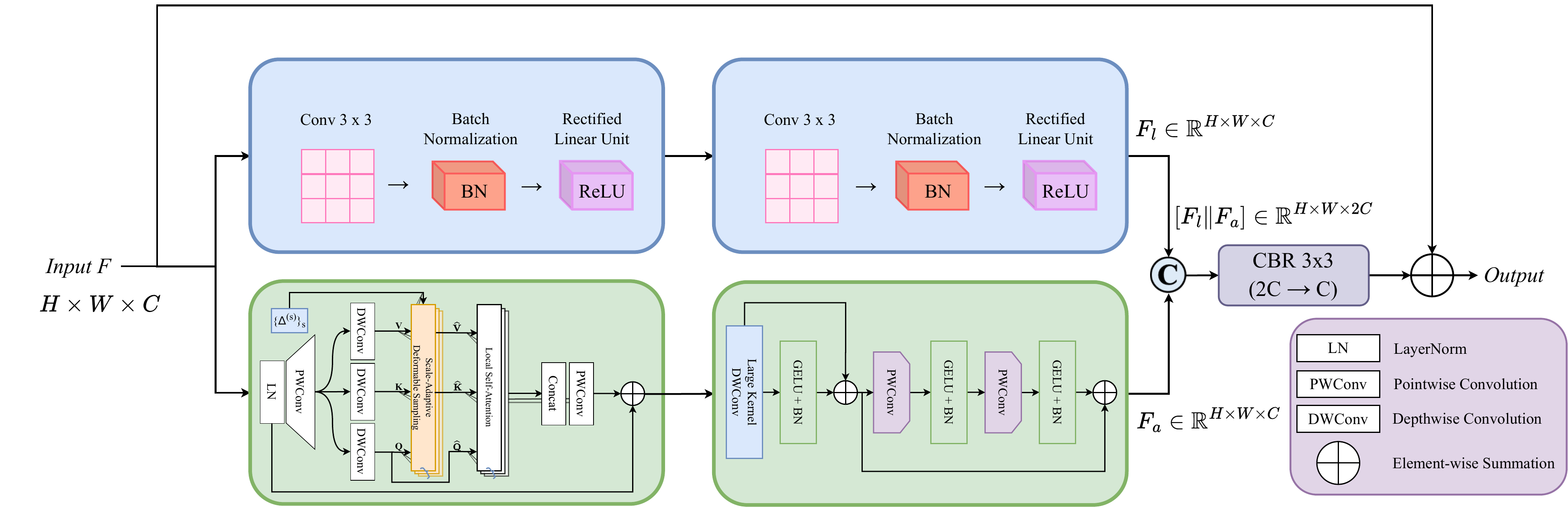}
\caption{Context-Guided Scale-Adaptive Deformable Encoding (CSDE). The upper branch extracts the fixed-grid feature $F_l$ using two $3\times3$ CBR blocks. The lower branch produces $F_a$ through deformable local attention, large-kernel depthwise convolution, and pointwise refinement. Its four attention heads use base spacings $\{1,3,5,7\}$ with nine offset-adjusted sampling locations per head. The branch outputs are concatenated, projected from $2C$ to $C$ channels, and added to the input. Circled C and $\oplus$ denote channel concatenation and element-wise addition, respectively; CBR denotes convolution, batch normalization, and ReLU.}
\label{fig:csde}
\end{figure*}

CSDE combines a fixed-grid branch with an adaptive branch (Fig.~\ref{fig:csde}). Two $3\times3$ CBR blocks preserve the feature shape and supply local information independently of the learned offsets. The adaptive branch uses learnable spatial support, following the principle of deformable operations \cite{Dai2017Deformable,Xia2022DAT}.

Channel normalization, pointwise projection, and depthwise refinement produce query, key, and value features. For a stage width $C$, the four heads each have width $d_h=C/4$. Head indices are $h\in\{1,2,3,4\}$, with corresponding base spacings $s_h\in\{1,3,5,7\}$. Let $r_j\in\{-1,0,1\}^2$, $j=1,\ldots,9$, enumerate a $3\times3$ reference lattice, and let $j_0$ denote its center. The sampling position is
\begin{equation}
 p_{h,j}=p+s_h r_j+\Delta_{h,j}(p),\quad j=1,\ldots,9,
\label{eq:sampling}
\end{equation}
where $p$ and the learned two-dimensional offsets are expressed in the coordinates of the current feature map. Here, $h$ identifies a channel group and $s_h$ determines its nominal spatial support.

Define $\mathcal B(U,x)$ as bilinear sampling of a feature map $U$ at location $x$, with border padding outside the feature grid. The sampled query is $\widehat q_h(p)=\mathcal B(Q_h,p+\Delta_{h,j_0}(p))$. Keys and values are sampled at all nine locations: $\widehat k_{h,j}(p)=\mathcal B(K_h,p_{h,j})$ and $\widehat v_{h,j}(p)=\mathcal B(V_h,p_{h,j})$. Each head computes
\begin{equation}
 O_h(p)=\sum_{j=1}^{9}\operatorname{softmax}_{j}\!\left(
 \frac{\widehat q_h(p)^\top\widehat k_{h,j}(p)}{\sqrt{d_h}}\right)\widehat v_{h,j}(p).
\label{eq:adaptive}
\end{equation}
The softmax is over the nine sampled locations for a fixed head and query. Head outputs are concatenated and projected to $C$ channels. A $7\times7$ depthwise convolution and a two-layer pointwise block then refine this output, using residual connections within the adaptive branch.

The two branch outputs, each of width $C$, are concatenated to width $2C$ and mapped back to $C$ by a $3\times3$ CBR block before addition to the input. The learned offsets adapt these neighborhoods to the input features, while the fixed-grid branch retains local convolutional support.

\subsection{Spatially adaptive multi-kernel gating}\label{sec:samg}
SAMG processes $E_s$ independently at stages $s\in\{2,3,4\}$. Its four dense CBR branches use kernel sizes in $\mathcal K=\{1,3,5,7\}$, preserving the spatial size and channel width of that stage. Their outputs, denoted by $B_{s,k}$, describe different neighborhoods on the same feature map rather than separate image resolutions.

\begin{figure}[pos=t]
\centering\includegraphics[width=\columnwidth]{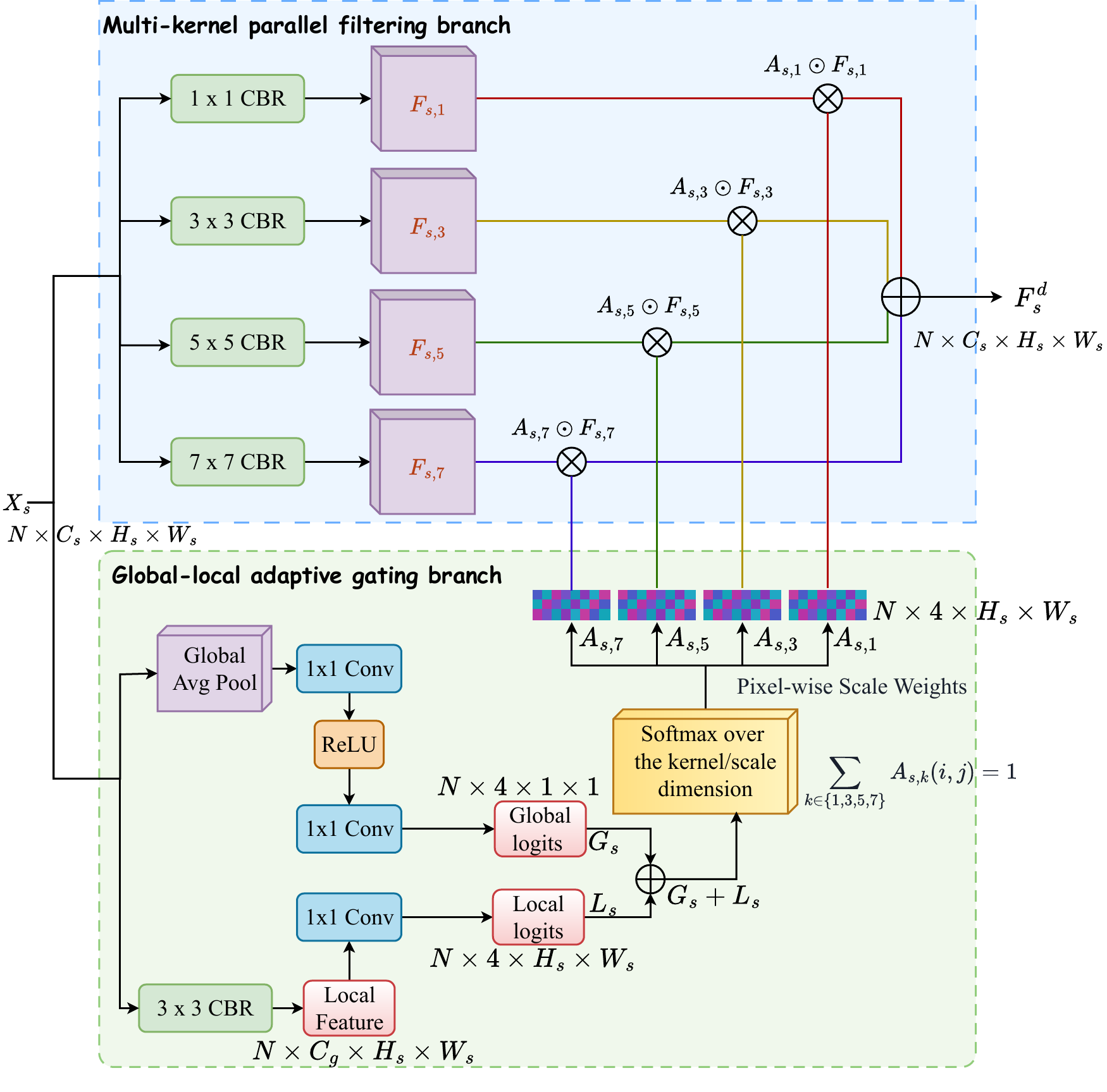}
\caption{Spatially Adaptive Multi-Kernel Gating (SAMG). Parallel CBR branches extract responses with kernel sizes $1$, $3$, $5$, and $7$. Pooled global logits are broadcast and added to local logits, then normalized by softmax across the four branches. The resulting weights vary by location and are shared across channels. Weighted summation produces $F_s^d$ without changing the stage resolution or channel width. Here, $\odot$ denotes channel-broadcast multiplication; CBR denotes convolution, batch normalization, and ReLU.}
\label{fig:samg}
\end{figure}

Two pathways predict branch logits (Fig.~\ref{fig:samg}). The global pathway applies average pooling and two pointwise convolutions to produce $G_s\in\mathbb R^4$. The local pathway uses a $3\times3$ transformation and a pointwise projection to produce $L_s\in\mathbb R^{4\times H_s\times W_s}$. The global logits are broadcast over the spatial grid. Their sum is normalized across branches:
\begin{align}
 A_{s,k}(p)&=\frac{\exp(G_{s,k}+L_{s,k}(p))}
 {\sum_{u\in\mathcal K}\exp(G_{s,u}+L_{s,u}(p))},\nonumber\\
 F_s^d(p)&=\sum_{k\in\mathcal K}A_{s,k}(p)B_{s,k}(p).
\label{eq:samg}
\end{align}
Thus $A_{s,k}(p)\geq0$ and the four weights sum to one. A branch weight is shared across feature channels at a location, but can differ between locations. The global pathway describes the current input patch, not the complete fundus image during patch-based training.

SAMG thus performs spatial selection over learned convolutional responses. The branch kernels remain fixed after training, while their contributions vary with the input at each location. The gated features retain their stage-specific channel widths until DCDF aligns and projects them.

\subsection{Dynamic cross-scale detail fusion}\label{sec:dcdf}
DCDF maps the three gated features to the $E_3$ resolution. The $E_2$ feature is downsampled by two, $E_3$ follows an identity alignment, and $E_4$ is bilinearly upsampled. Independent $1\times1$ CBR projections reduce each aligned feature to eight channels. With $\Proj_s$ denoting alignment followed by projection, the fused feature is
\begin{equation}
 F_{\mathrm{detail}}=\CBR_1\!\left[
 \Proj_2(F_2^d)\Vert\Proj_3(F_3^d)\Vert\Proj_4(F_4^d)\right].
\label{eq:detail}
\end{equation}
The concatenated tensor has 24 channels, and the final CBR projection returns eight. The output lies at nominal quarter input resolution.

\begin{figure}[pos=t]
\centering\includegraphics[width=.86\columnwidth]{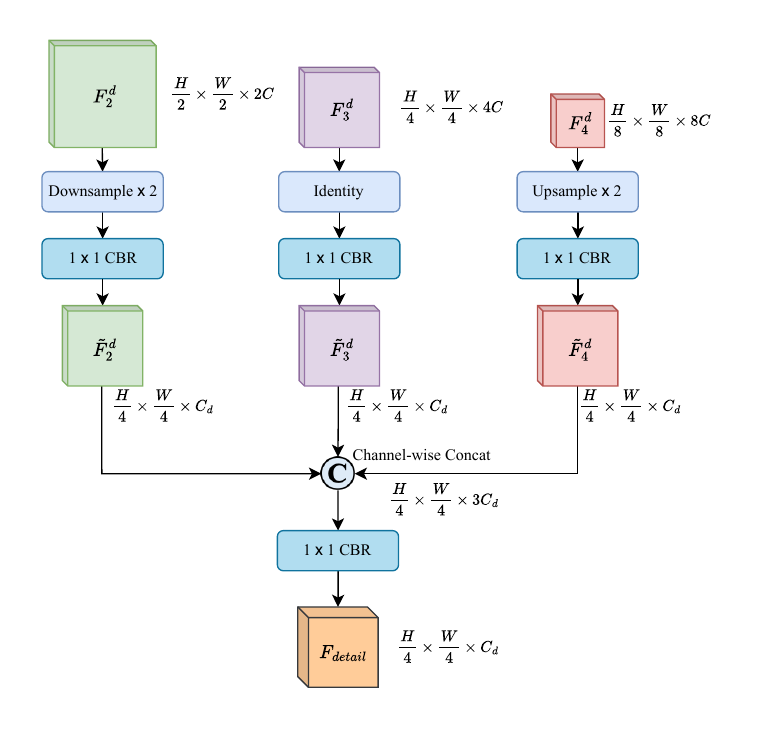}
\caption{Dynamic Cross-Scale Detail Fusion (DCDF). Gated features $F_2^d$, $F_3^d$, and $F_4^d$ are aligned to the $E_3$ grid by downsampling, identity mapping, and upsampling, respectively. Independent $1\times1$ CBR projections map each feature to $C_d=8$ channels. Concatenation followed by a $3C_d$-to-$C_d$ projection produces the shared feature $F_{\mathrm{detail}}$ at quarter input resolution. CBR denotes convolution, batch normalization, and ReLU.}
\label{fig:dcdf}
\end{figure}

DCDF combines the input-adaptive responses from SAMG through spatial alignment and channel fusion. Its eight-channel output gives successive decoder stages access to the same multi-scale information through a compact intermediate representation.

\subsection{Detail-guided decoding and selective skips}\label{sec:decoder}
Each decoder block contains two $3\times3$ CBR transformations and spatial dropout with rate 0.1. Let $\mathcal U_i$ denote bilinear resizing to the target stage followed by a $1\times1$ CBR channel projection. Let $\Phi_i$ denote the fusion transformation after concatenation, and $\Dec_i$ the decoder block. The deepest feature initializes $D_5=\Dec_5(E_5)$; the next stage fuses $\mathcal U_4(D_5)$ with $E_4$ to obtain $D_4$.

For $i\in\{3,2,1\}$, write $T_i=\Psi_i(\mathcal R_i(F_{\mathrm{detail}}))$, where $\mathcal R_i$ resizes to the $D_i$ grid and $\Psi_i$ is the stage-specific $1\times1$ CBR projection. The channel widths of $T_3$, $T_2$, and $T_1$ are 32, 16, and 8. Although the projections are learned independently, they all receive the same cross-scale fusion output. The remaining stages are
\begin{align}
 D_3 &= \Dec_3\!\left(\CCA\!\left(\Phi_3[\mathcal{U}_3(D_4)\Vert T_3]\right)\right),\nonumber\\
 D_2 &= \Dec_2\!\left(\Phi_2[\mathcal{U}_2(D_3)\Vert T_2]\right),\nonumber\\
 D_1 &= \Dec_1\!\left(\Phi_1[\mathcal{U}_1(D_2)\Vert E_1\Vert T_1]\right).
\label{eq:decode}
\end{align}
A single criss-cross attention (CCA) block \cite{Huang2020CCNet} refines D3 fusion along rows and columns. Direct skips from $E_2$ and $E_3$ are omitted, whereas $E_1$ is retained at the final stage. A bare $1\times1$ convolution maps $D_1$ to one-channel logits. Resizing uses the exact spatial size of the target feature tensor, including for odd-sized inputs.

\subsection{Training objective}
Let $z_i$ be a vessel logit, $y_i\in\{0,1\}$ its reference label, and $M_i\in\{0,1\}$ the FOV mask. The objective is
\begin{equation}
 \mathcal{L}=-\frac{\sum_i M_i\left[y_i\log\sigma(z_i)+(1-y_i)\log(1-\sigma(z_i))\right]}
 {\sum_i M_i}.
\label{eq:objective}
\end{equation}
The sum includes the batch and spatial dimensions of the training patches. Only FOV pixels contribute, and their count in the denominator must be positive. Binary cross-entropy is evaluated directly from logits for numerical stability; every reported configuration uses this same objective without additional loss terms.

\section{Experimental results}\label{sec:experiments}
We evaluate in-domain segmentation on four datasets, examine the architectural construction through cumulative ablations, and test all twelve directed transfers between datasets. Literature-based numerical comparisons and locally reproduced qualitative predictions are identified separately.

\subsection{Datasets}
Experiments use DRIVE~\cite{Staal2004DRIVE}, CHASE\_DB1~\cite{Fraz2012CHASE}, STARE~\cite{Hoover2000STARE}, and HRF~\cite{Budai2013HRF}. Table~\ref{tab:datasets} summarizes their image counts, resolutions, and evaluation divisions.

\begin{table}[pos=t]
\caption{An overview of our datasets.}
\label{tab:datasets}
\centering
\fontsize{7.8}{9.4}\selectfont
\renewcommand{\arraystretch}{1.15}
\setlength{\tabcolsep}{2.0pt}
\begin{tabular*}{\columnwidth}{@{\extracolsep{\fill}}lccc}
\toprule
Dataset & Quantity & Resolution & Train - Test \\
\midrule
DRIVE & 40 & $565\times584$ & 20 - 20 \\
CHASE\_DB1 & 28 & $999\times960$ & 20 - 8 \\
STARE & 20 & $700\times605$ & 5 - Fold \\
HRF & 45 & $3504\times2336$ & 25 - 20 \\
\bottomrule
\end{tabular*}
\end{table}

\textit{DRIVE.} DRIVE contains 40 images at $565\times584$ pixels. We use the official division, with images 21--40 for training and 01--20 for testing, giving 20 images in each partition.

\textit{CHASE\_DB1.} CHASE\_DB1 contains 28 images from both eyes of 14 subjects at $999\times960$ pixels. The reported experiments use 20 training images and eight test images. \textit{STARE.} STARE contains 20 images at $700\times605$ pixels, including normal and pathological appearances. We use five-fold evaluation because the dataset has no official training-test partition.

\textit{HRF.} HRF contains 45 images at $3504\times2336$ pixels, covering healthy subjects, diabetic retinopathy, and glaucoma \cite{Budai2013HRF,HRFOfficial}. The experiments use 25 training images and 20 test images.

\subsection{Evaluation metrics}\label{sec:metrics}
We report sensitivity (SE), specificity (SP), accuracy (ACC), F1, and the area under the receiver operating characteristic curve (AUC), as defined in Table~\ref{tab:metrics}. Vessel pixels are the positive class. SE and SP describe vessel detection and background rejection, whereas F1 measures foreground overlap and AUC evaluates discrimination across thresholds. Together, they characterize both the probability map and its binary segmentation.

\begin{table}[pos=t]
\caption{Metrics for evaluating the performance.}
\label{tab:metrics}
\centering
\fontsize{7.8}{9.4}\selectfont
\renewcommand{\arraystretch}{1.18}
\setlength{\tabcolsep}{2.0pt}
\begin{tabular*}{\columnwidth}{@{\extracolsep{\fill}}ll@{}}
\toprule
Metrics & Calculation formula \\
\midrule
SE (Sensitivity) & $\mathrm{SE}=\mathrm{TP}/(\mathrm{TP}+\mathrm{FN})$ \\
SP (Specificity) & $\mathrm{SP}=\mathrm{TN}/(\mathrm{TN}+\mathrm{FP})$ \\
ACC (Accuracy) & $\mathrm{ACC}=(\mathrm{TP}+\mathrm{TN})/(\mathrm{TP}+\mathrm{TN}+\mathrm{FP}+\mathrm{FN})$ \\
F1 (F1 score) & $\mathrm{F1}=(2\times\mathrm{TP})/(2\times\mathrm{TP}+\mathrm{FP}+\mathrm{FN})$ \\
AUC & Area Under the ROC Curve. \\
\bottomrule
\end{tabular*}
\end{table}

Binary predictions use a fixed probability threshold of 0.5. Metrics are computed separately for each image and then averaged without weighting by image size. An available FOV mask restricts evaluation to retinal pixels; otherwise, the full image is used. AUC is calculated from the reconstructed probability map before thresholding. F1 equals binary Dice when the masks and evaluation pixels are identical. The image-wise averaging used here is distinct from pooling pixels across images \cite{Galdran2022Minimalistic}.

Segmentation scores in Tables~\ref{tab:drive_comparison}--\ref{tab:ablation_drive} are fractions. Table~\ref{tab:cross_validation} gives percentages and percentage-point differences, and the AUC and F1 columns in Table~\ref{tab:complexity} also use percentages. Table~\ref{tab:statistical_analysis} presents AUC distribution summaries and the reported statistical comparisons.

\subsection{Implementation details}\label{sec:implementation}
\textit{Experimental environment.} The recorded experimental configuration specifies an NVIDIA H200 GPU with 141 GB memory, PyTorch 2.13.0, and CUDA 13.2. CGDD-Net contains 1,956,802 trainable parameters, with encoder widths $8,16,32,64,128$ and an eight-channel shared detail representation.

\textit{Optimization.} Adam~\cite{Kingma2015Adam} uses a base learning rate of 0.001, zero weight decay, batch size 64, and a maximum of 100 epochs. A linear warmup increases the learning rate from 0.1 of its base value during the first five epochs. Cosine annealing with warm restarts then uses an initial 20-epoch cycle and doubles the subsequent cycle lengths \cite{Loshchilov2017SGDR}. The default random seed is 42.

\textit{Patch sampling and augmentation.} Each epoch samples 150,000 training patches of $64\times64$ pixels. Images are converted to grayscale and scaled to $[0,1]$, and patch centers are sampled inside the retinal FOV. Horizontal and vertical flips with probability 0.5 and rotations by multiples of $90^{\circ}$ are applied jointly to each image, label, and FOV mask. Decoder blocks use spatial dropout with rate 0.1.

\textit{Inference and reconstruction.} Inference uses $96\times96$ windows with stride 16 and batch size eight. Reflection padding on the right and bottom provides sliding-window coverage. For pixel $p$, the reconstructed probability is $P(p)=|\Omega(p)|^{-1}\sum_{w\in\Omega(p)}\sigma(z_w(p))$, where $\Omega(p)$ is the set of windows covering $p$ and $z_w(p)$ is its logit in window $w$. Averaging is performed after the sigmoid operation.

The map is then cropped to the original image size and thresholded at 0.5. No morphological post-processing is applied.

\textit{Computational profile.} Table~\ref{tab:complexity} lists 1.96M parameters and a reported model-forward count of 29.96 GFLOPs. Model-forward counts are distinguished from the image-level reconstruction pipeline, which also includes window extraction, probability transfer, and overlap aggregation. The table retains source-reported FLOPs for descriptive comparison rather than a measurement of end-to-end latency under a common profiling configuration.

\textit{Qualitative baselines.} U-Net3+~\cite{Huang2020UNet3Plus}, U-Net~\cite{Ronneberger2015UNet}, LadderNet~\cite{Zhuang2019LadderNet}, ConvUNeXt~\cite{Han2022ConvUNeXt}, and AG-Net~\cite{Zhang2019AGNet} are reproduced from public implementations or published architectures for Fig.~\ref{fig:qualitative}. These predictions use the same stated dataset partitions, preprocessing, augmentation, reconstruction, FOV handling, and threshold as CGDD-Net. The numerical results in the literature comparison tables remain those reported by their respective sources.

\subsection{Comparison with existing methods}\label{sec:comparison}
Tables~\ref{tab:drive_comparison}--\ref{tab:hrf_comparison} place CGDD-Net alongside existing retinal vessel segmentation methods. Literature values follow the original studies' data divisions, annotations, and evaluation protocols. Red and cyan denote numerical order within each displayed comparison set; our analysis identifies the metrics on which CGDD-Net performs particularly well.

\begin{table}[pos=htbp]
\caption{Comparison on DRIVE. Best and second-best values per metric are highlighted in red and cyan, respectively; literature results follow the protocols of the original studies.}
\label{tab:drive_comparison}
\centering
\fontsize{6.2}{7.0}\selectfont\renewcommand{\arraystretch}{1.00}
\setlength{\tabcolsep}{0.35pt}
\begin{tabular*}{\linewidth}{@{\extracolsep{\fill}}lccccc}
\toprule
\CGDDTableHeadStrut Methods & SE & SP & ACC & F1 & AUC \\
\midrule
U-Net \cite{Ronneberger2015UNet} & 0.7537 & \textcolor{cyan}{0.9820} & 0.9531 & 0.8142 & 0.9755 \\
LadderNet \cite{Zhuang2019LadderNet} & 0.7856 & 0.9810 & 0.9561 & 0.8202 & 0.9793 \\
CE-Net \cite{Gu2019CENet} & 0.8209 & 0.9751 & 0.9547 & 0.8202 & 0.9795 \\
DEU-Net \cite{Wang2019DEUNet} & 0.7940 & 0.9816 & 0.9567 & 0.8270 & 0.9772 \\
Xu \cite{Xu2021Joint} & 0.8339 & 0.9750 & 0.9571 & 0.8319 & 0.9821 \\
MAU-Net \cite{Li2020MAUNet} & 0.7890 & 0.9799 & 0.9557 & 0.8192 & 0.9774 \\
CS$^{2}$-Net \cite{Mou2021CS2} & 0.8154 & 0.9757 & 0.9553 & 0.8228 & 0.9784 \\
MPS-Net \cite{Lin2021MPSNet} & 0.8361 & 0.9740 & 0.9563 & 0.8287 & 0.9805 \\
MFI-Net \cite{Ye2022MFINet} & 0.8170 & 0.9790 & 0.9581 & 0.8315 & 0.9836 \\
AACA-MLA-D-U-Net \cite{Yuan2022MLANet} & 0.8046 & 0.9805 & 0.9581 & \textcolor{red}{0.8530} & 0.9827 \\
DPF-Net \cite{Li2023DPFNet} & 0.8279 & 0.9776 & 0.9570 & 0.8303 & 0.9824 \\
AMFEF-Net \cite{Zhu2025AMFEF} & 0.8084 & 0.9802 & 0.9583 & 0.8315 & 0.9827 \\
MSVENet \cite{Cao2025MSVENet} & \textcolor{red}{0.8432} & 0.9799 & 0.9631 & \textcolor{cyan}{0.8487} & \textcolor{cyan}{0.9853} \\
FS-Net \cite{Getahun2023FSNet} & \textcolor{cyan}{0.8421} & \textcolor{red}{0.9828} & \textcolor{cyan}{0.9702} & 0.8314 & \textcolor{red}{0.9884} \\
AttUKAN \cite{Zeng2025AttUKAN} & 0.8393 & 0.9721 & 0.9549 & 0.8250 & 0.9795 \\
\midrule
CGDD-Net (Ours) & \textcolor{cyan}{0.8421} & 0.9768 & \textcolor{red}{0.9704} & 0.8323 & 0.9824 \\
\bottomrule
\end{tabular*}
\end{table}

\textit{DRIVE.} CGDD-Net achieves SE 0.8421, SP 0.9768, ACC 0.9704, F1 0.8323, and AUC 0.9824. It has the highest listed ACC, exceeding FS-Net by 0.02 percentage points, and shares the second-highest SE with FS-Net \cite{Getahun2023FSNet}. Its F1 is higher than the listed results of LadderNet, CE-Net, MPS-Net, and MFI-Net. The combination of vessel recall and foreground overlap demonstrates the effectiveness of the complete architecture on this benchmark.

\begin{table}[pos=htbp]
\caption{Comparison on CHASE\_DB1. Best and second-best values per metric are highlighted in red and cyan, respectively; literature results follow the protocols of the original studies.}
\label{tab:chase_comparison}
\centering
\fontsize{6.2}{7.0}\selectfont\renewcommand{\arraystretch}{1.00}
\setlength{\tabcolsep}{0.35pt}
\begin{tabular*}{\linewidth}{@{\extracolsep{\fill}}lccccc}
\toprule
\CGDDTableHeadStrut Methods & SE & SP & ACC & F1 & AUC \\
\midrule
U-Net \cite{Ronneberger2015UNet} & 0.8288 & 0.9701 & 0.9578 & 0.7783 & 0.9772 \\
LadderNet \cite{Zhuang2019LadderNet} & 0.7978 & 0.9818 & 0.9656 & 0.8031 & 0.9839 \\
CE-Net \cite{Gu2019CENet} & 0.8027 & 0.9813 & 0.9651 & 0.8012 & 0.9827 \\
DEU-Net \cite{Wang2019DEUNet} & 0.8074 & 0.9821 & 0.9661 & 0.8037 & 0.9812 \\
Xu \cite{Xu2021Joint} & \textcolor{cyan}{0.8508} & 0.9780 & 0.9651 & 0.8270 & 0.9873 \\
Iter-Net \cite{Li2020IterNet} & 0.7970 & 0.9823 & 0.9655 & 0.8073 & 0.9851 \\
CS$^{2}$-Net \cite{Mou2021CS2} & 0.8329 & 0.9784 & 0.9651 & 0.8141 & 0.9851 \\
MPS-Net \cite{Lin2021MPSNet} & 0.8488 & 0.9795 & 0.9668 & \textcolor{red}{0.8332} & 0.9869 \\
MFI-Net \cite{Ye2022MFINet} & 0.8388 & 0.9806 & 0.9675 & 0.8205 & 0.9879 \\
AACA-MLA-D-U-Net \cite{Yuan2022MLANet} & 0.8402 & 0.9801 & 0.9673 & 0.8248 & 0.9874 \\
DPF-Net \cite{Li2023DPFNet} & 0.8303 & \textcolor{cyan}{0.9841} & 0.9676 & \textcolor{cyan}{0.8302} & 0.9868 \\
TA-Net \cite{Cui2025TANet} & 0.8268 & 0.9771 & 0.9635 & 0.8039 & 0.9829 \\
MSVENet \cite{Cao2025MSVENet} & 0.8200 & 0.9811 & 0.9665 & 0.8160 & 0.9860 \\
FS-Net \cite{Getahun2023FSNet} & 0.8365 & \textcolor{red}{0.9850} & \textcolor{red}{0.9755} & 0.8144 & \textcolor{cyan}{0.9903} \\
AttUKAN \cite{Zeng2025AttUKAN} & 0.8061 & 0.9811 & 0.9631 & 0.8134 & 0.9821 \\
\midrule
CGDD-Net (Ours) & \textcolor{red}{0.8600} & 0.9815 & \textcolor{cyan}{0.9752} & 0.8102 & \textcolor{red}{0.9938} \\
\bottomrule
\end{tabular*}
\end{table}

\textit{CHASE\_DB1.} CGDD-Net attains the highest listed AUC (0.9938) and SE (0.8600); ACC, F1, and SP are 0.9752, 0.8102, and 0.9815, respectively. Its AUC exceeds FS-Net's 0.9903 by 0.35 percentage points, emphasizing probability discrimination alongside the fixed-threshold F1 result.

\begin{table}[pos=htbp]
\caption{Comparison on STARE. Best and second-best values per metric are highlighted in red and cyan, respectively; literature results follow the protocols of the original studies.}
\label{tab:stare_comparison}
\centering
\fontsize{6.2}{7.0}\selectfont\renewcommand{\arraystretch}{1.00}
\setlength{\tabcolsep}{0.35pt}
\begin{tabular*}{\linewidth}{@{\extracolsep{\fill}}lccccc}
\toprule
\CGDDTableHeadStrut Methods & SE & SP & ACC & F1 & AUC \\
\midrule
U-Net \cite{Ronneberger2015UNet} & 0.6749 & \textcolor{cyan}{0.9886} & 0.9576 & 0.7655 & 0.9737 \\
LadderNet \cite{Zhuang2019LadderNet} & 0.7010 & \textcolor{red}{0.9895} & 0.9598 & 0.7817 & 0.9809 \\
CE-Net \cite{Gu2019CENet} & 0.8296 & 0.9841 & 0.9667 & 0.8296 & 0.9867 \\
HANet \cite{Wang2020HANet} & 0.8186 & 0.9844 & 0.9673 & 0.8379 & 0.9881 \\
Iter-Net \cite{Li2020IterNet} & 0.7715 & \textcolor{cyan}{0.9886} & 0.9701 & 0.8146 & 0.9881 \\
CS$^{2}$-Net \cite{Mou2021CS2} & 0.8396 & 0.9813 & 0.9670 & 0.8420 & 0.9875 \\
MPS-Net \cite{Lin2021MPSNet} & \textcolor{cyan}{0.8566} & 0.9819 & 0.9689 & 0.8491 & 0.9873 \\
MFI-Net \cite{Ye2022MFINet} & 0.8194 & 0.9806 & 0.9678 & 0.8320 & 0.9887 \\
AACA-MLA-D-U-Net \cite{Yuan2022MLANet} & 0.7914 & 0.9870 & 0.9665 & 0.8276 & 0.9864 \\
DPF-Net \cite{Li2023DPFNet} & 0.8287 & 0.9854 & 0.9655 & 0.8366 & 0.9854 \\
TA-Net \cite{Cui2025TANet} & 0.7280 & 0.9845 & 0.9582 & 0.7812 & 0.9793 \\
AMFEF-Net \cite{Zhu2025AMFEF} & 0.7307 & \textcolor{cyan}{0.9886} & 0.9622 & 0.7985 & 0.9839 \\
MSVENet \cite{Cao2025MSVENet} & 0.8504 & 0.9859 & 0.9703 & \textcolor{red}{0.8606} & 0.9893 \\
FS-Net \cite{Getahun2023FSNet} & 0.8393 & 0.9852 & \textcolor{cyan}{0.9750} & 0.8261 & \textcolor{red}{0.9916} \\
AttUKAN \cite{Zeng2025AttUKAN} & 0.7688 & 0.9868 & 0.9643 & 0.8114 & 0.9796 \\
\midrule
CGDD-Net (Ours) & \textcolor{red}{0.8661} & 0.9812 & \textcolor{red}{0.9775} & \textcolor{cyan}{0.8510} & \textcolor{cyan}{0.9895} \\
\bottomrule
\end{tabular*}
\end{table}

\textit{STARE.} CGDD-Net reaches SE 0.8661 and ACC 0.9775, the highest values in Table~\ref{tab:stare_comparison}. Its F1 of 0.8510 and AUC of 0.9895 rank second within that table. The SE and ACC gains over the next listed values are 0.95 and 0.25 percentage points, respectively. The model-size comparison in Table~\ref{tab:complexity} additionally includes MDF-Net, providing a separate performance-cost comparison.

\begin{table}[pos=htbp]
\caption{Comparison with existing methods on HRF. Best and second-best values in each metric column are highlighted in red and cyan, respectively; literature results follow the evaluation protocols reported by the original studies. A dash denotes an unreported metric, and the M2U-Net F1 entry is its reported binary Dice score.}
\label{tab:hrf_comparison}
\centering
\CGDDTableStyle
\setlength{\tabcolsep}{1.5pt}
\begin{tabular*}{\linewidth}{@{\extracolsep{\fill}}lccccc}
\toprule
\CGDDTableHeadStrut Method & SE & SP & ACC & F1 & AUC \\
\midrule
U-Net \cite{Ronneberger2015UNet} & 0.7794 & 0.9804 & 0.9641 & 0.7967 & 0.9721 \\
DUNet \cite{Jin2019DUNet} & 0.7464 & \textcolor{cyan}{0.9874} & 0.9651 & -- & 0.9831 \\
HANet \cite{Wang2020HANet} & 0.7803 & 0.9843 & 0.9653 & \textcolor{cyan}{0.8074} & 0.9837 \\
MFI-Net \cite{Ye2022MFINet} & 0.7544 & \textcolor{red}{0.9884} & \textcolor{cyan}{0.9668} & 0.8054 & \textcolor{cyan}{0.9858} \\
M2U-Net \cite{Laibacher2018M2UNet} & -- & -- & 0.9635 & 0.7814 & -- \\
DPN \cite{Guo2020DPN} & \textcolor{cyan}{0.7926} & 0.9764 & 0.9591 & 0.7835 & 0.9697 \\
AttUKAN \cite{Zeng2025AttUKAN} & 0.7902 & 0.9817 & 0.9645 & 0.8021 & 0.9789 \\
CGDD-Net (Ours) & \textcolor{red}{0.8362} & 0.9823 & \textcolor{red}{0.9711} & \textcolor{red}{0.8157} & \textcolor{red}{0.9874} \\
\bottomrule
\end{tabular*}
\end{table}

\textit{HRF.} CGDD-Net achieves the highest listed SE (0.8362), ACC (0.9711), F1 (0.8157), and AUC (0.9874). F1 exceeds HANet~\cite{Wang2020HANet} by 0.83 percentage points, and AUC exceeds MFI-Net~\cite{Ye2022MFINet} by 0.16 points. These results extend the evaluation to native high-resolution fundus images and support the use of the shared detail architecture across different image dimensions.

\begin{figure*}[pos=p]
\centering\includegraphics[width=.98\textwidth,height=.83\textheight,keepaspectratio]{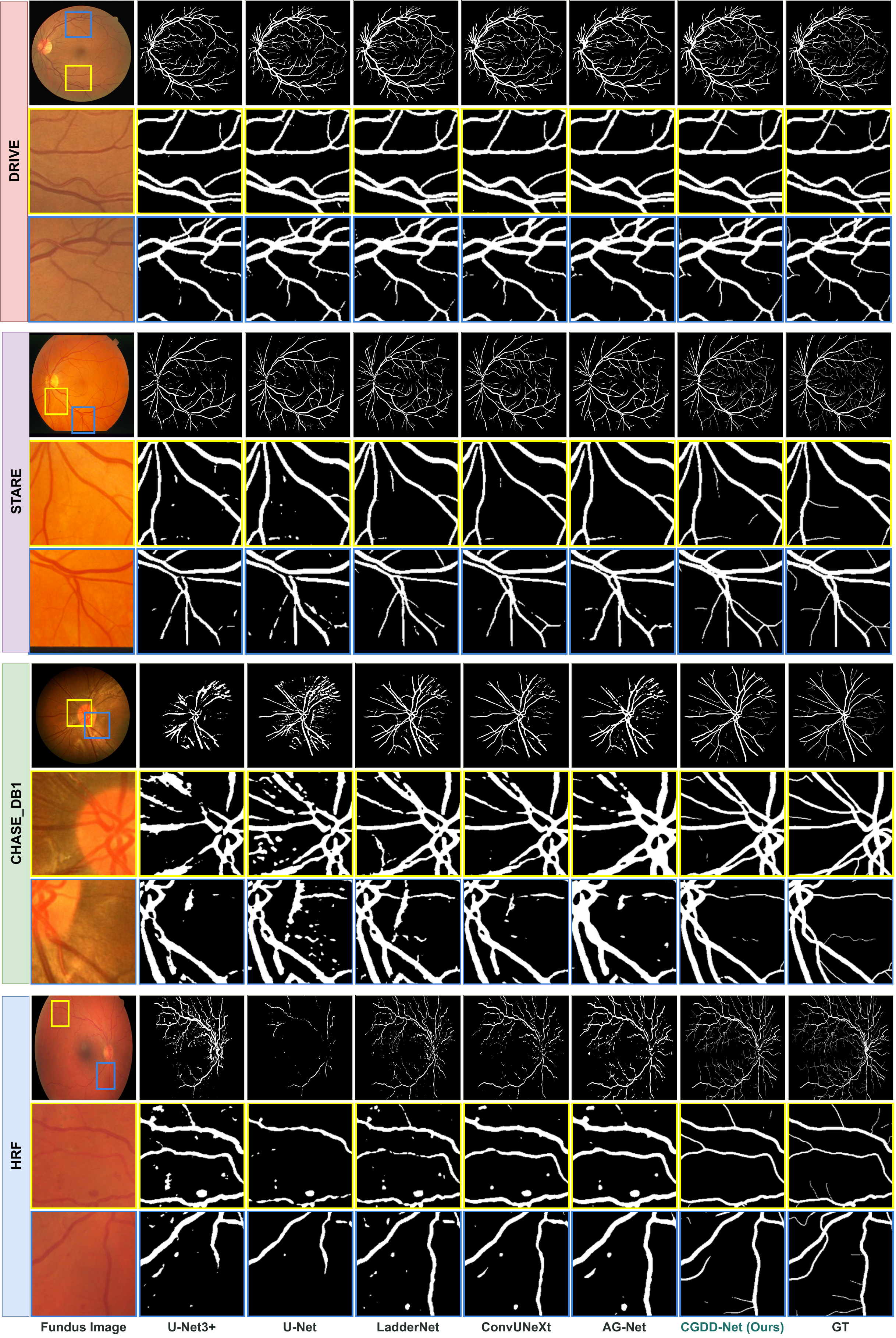}
\caption{Qualitative vessel-segmentation comparison. Dataset groups are ordered from top to bottom as DRIVE, STARE, CHASE\_DB1, and HRF. Columns show the fundus image, U-Net3+, U-Net, LadderNet, ConvUNeXt, AG-Net, CGDD-Net, and ground truth (GT). Each group contains a full-image view followed by matched enlargements of the yellow- and blue-boxed regions. The crops allow direct comparison of narrow branches, vessel boundaries, and separation between adjacent vessels.}
\label{fig:qualitative}
\end{figure*}

\textit{Qualitative comparison.}\label{sec:visual} Fig.~\ref{fig:qualitative} presents full-image predictions and enlarged regions for the reproduced baselines and CGDD-Net. The crops highlight branch continuity, boundary delineation, and isolated responses. In several highlighted regions, CGDD-Net preserves narrow connections that appear fragmented in comparison outputs. The full-image views place these local differences in the context of the reconstructed vascular tree.

\textit{AUC distribution summaries.}\label{sec:statistics} Table~\ref{tab:statistical_analysis} reports AUC means, standard deviations, and the reported Wilcoxon rank-sum comparisons. CGDD-Net has the highest mean AUC among the methods listed for CHASE\_DB1 and STARE, while its DRIVE mean is close to those of MFI-Net and AACA-MLA-D-U-Net. The source summaries do not specify a common sampling unit, so the reported $p$-values are retained for reference and are not used to establish confirmatory significance. The interpretation here rests on the observed AUC estimates and the segmentation results above.

\begin{table*}[t]
\caption{AUC summaries and reported Wilcoxon rank-sum comparisons on DRIVE, CHASE\_DB1, and STARE. AUC is given as mean $\pm$ standard deviation; $^{*}$ denotes a reported $p<0.05$ for comparison with CGDD-Net. Red and cyan identify the two highest means in each dataset. Section~\ref{sec:statistics} states the descriptive use of these summaries.}
\label{tab:statistical_analysis}
\centering
\CGDDTableStyle
\setlength{\tabcolsep}{4pt}
\begin{tabular*}{\textwidth}{@{\extracolsep{\fill}}lcccccc}
\toprule
Method & DRIVE AUC & $p$-value & CHASE\_DB1 AUC & $p$-value & STARE AUC & $p$-value \\
\midrule
U-Net \cite{Ronneberger2015UNet}
& 0.9755 $\pm$ 0.0008 & $5.26\times10^{-9}$\textsuperscript{*}
& 0.9772 $\pm$ 0.0010 & $1.51\times10^{-9}$\textsuperscript{*}
& 0.9737 $\pm$ 0.0006 & $4.76\times10^{-23}$\textsuperscript{*} \\
LadderNet \cite{Zhuang2019LadderNet}
& 0.9793 $\pm$ 0.0014 & $1.64\times10^{-2}$\textsuperscript{*}
& 0.9839 $\pm$ 0.0010 & $5.45\times10^{-5}$\textsuperscript{*}
& 0.9809 $\pm$ 0.0009 & $2.81\times10^{-9}$\textsuperscript{*} \\
CE-Net \cite{Gu2019CENet}
& 0.9795 $\pm$ 0.0013 & $1.98\times10^{-2}$\textsuperscript{*}
& 0.9827 $\pm$ 0.0018 & $1.32\times10^{-6}$\textsuperscript{*}
& 0.9867 $\pm$ 0.0010 & $4.32\times10^{-2}$\textsuperscript{*} \\
MPS-Net \cite{Lin2021MPSNet}
& 0.9805 $\pm$ 0.0014 & $5.43\times10^{-2}$
& 0.9869 $\pm$ 0.0009 & $3.50\times10^{-2}$\textsuperscript{*}
& 0.9873 $\pm$ 0.0012 & $1.41\times10^{-2}$\textsuperscript{*} \\
MFI-Net \cite{Ye2022MFINet}
& \textcolor{red}{0.9836 $\pm$ 0.0004} & $7.75\times10^{-2}$
& \textcolor{cyan}{0.9879 $\pm$ 0.0005} & $5.55\times10^{-1}$
& \textcolor{cyan}{0.9887 $\pm$ 0.0007} & $7.90\times10^{-3}$\textsuperscript{*} \\
CS$^{2}$-Net \cite{Mou2021CS2}
& 0.9784 $\pm$ 0.0007 & $2.09\times10^{-5}$\textsuperscript{*}
& 0.9851 $\pm$ 0.0008 & $4.79\times10^{-4}$\textsuperscript{*}
& 0.9875 $\pm$ 0.0012 & $1.56\times10^{-4}$\textsuperscript{*} \\
AACA-MLA-D-U-Net \cite{Yuan2022MLANet}
& \textcolor{cyan}{0.9827 $\pm$ 0.0009} & $1.13\times10^{-2}$\textsuperscript{*}
& 0.9874 $\pm$ 0.0007 & $9.56\times10^{-4}$\textsuperscript{*}
& 0.9864 $\pm$ 0.0011 & $7.86\times10^{-4}$\textsuperscript{*} \\
\midrule
\textbf{CGDD-Net (Ours)}
& \textbf{0.9824 $\pm$ 0.0192} & --
& \textcolor{red}{\textbf{0.9938 $\pm$ 0.0048}} & --
& \textcolor{red}{\textbf{0.9895 $\pm$ 0.0163}} & -- \\
\bottomrule
\end{tabular*}
\end{table*}

\subsection{Ablation studies}\label{sec:ablation}
Table~\ref{tab:ablation_drive} evaluates the construction of CGDD-Net on DRIVE, CHASE\_DB1, and STARE. The internal baseline uses convolutional encoder blocks, a standard decoder, and direct skips. Successive rows introduce CSDE, SAMG, DCDF, detail-guided decoding, and selective skips. The final row adds CCA after D3 fusion. Each row extends the preceding configuration, giving a cumulative comparison under the same segmentation objective.

\begin{table*}[pos=t]
\caption{Cumulative ablation study on DRIVE, CHASE\_DB1, and STARE. Each row builds on the preceding configuration, following the progression described in Section~\ref{sec:ablation}. The final CGDD-Net row additionally includes the CCA refinement after D3 fusion. The best and second-best values in each metric column are highlighted in red and cyan, respectively.}
\label{tab:ablation_drive}
\centering
\CGDDTableStyle
\setlength{\tabcolsep}{0.75pt}
\begin{tabular*}{\textwidth}{@{\extracolsep{\fill}}lccccccccccccccc@{}}
\toprule
\CGDDTableHeadStrut \multirow{2}{*}[-2.5pt]{Model} & \multicolumn{5}{c}{DRIVE} & \multicolumn{5}{c}{CHASE\_DB1} & \multicolumn{5}{c}{STARE}\\
\cmidrule(lr){2-6}\cmidrule(lr){7-11}\cmidrule(l){12-16}
\CGDDTableSubHeadStrut & SE & SP & ACC & F1 & AUC & SE & SP & ACC & F1 & AUC & SE & SP & ACC & F1 & AUC\\
\midrule
Baseline & 0.7537 & 0.9820 & 0.9531 & 0.8075 & 0.9801 & 0.8063 & 0.9820 & 0.9607 & 0.8041 & 0.9817 & 0.7810 & 0.9778 & 0.9630 & 0.8161 & 0.9684\\
+ CSDE & 0.7974 & 0.9822 & 0.9672 & 0.8139 & 0.9805 & 0.8220 & 0.9824 & 0.9655 & 0.8059 & 0.9821 & 0.7965 & 0.9802 & 0.9655 & 0.8271 & 0.9729\\
+ SAMG & 0.8157 & \textcolor{red}{0.9824} & 0.9681 & 0.8201 & 0.9816 & 0.8527 & 0.9789 & 0.9713 & 0.8063 & 0.9839 & 0.8431 & 0.9731 & 0.9669 & 0.8373 & 0.9817\\
+ DCDF & 0.8227 & \textcolor{cyan}{0.9823} & 0.9688 & 0.8258 & 0.9821 & 0.8466 & 0.9859 & 0.9725 & 0.8086 & 0.9870 & 0.8462 & 0.9817 & 0.9695 & 0.8363 & 0.9859\\
+ Detail-Guided Decoder & 0.8265 & 0.9818 & 0.9697 & 0.8301 & \textcolor{cyan}{0.9823} & 0.8532 & \textcolor{cyan}{0.9877} & 0.9734 & \textcolor{red}{0.8197} & 0.9887 & 0.8470 & \textcolor{cyan}{0.9860} & 0.9720 & 0.8424 & 0.9867\\
+ Selective Skip & \textcolor{cyan}{0.8373} & 0.9819 & \textcolor{cyan}{0.9700} & \textcolor{cyan}{0.8314} & 0.9822 & \textcolor{cyan}{0.8558} & \textcolor{red}{0.9906} & \textcolor{cyan}{0.9749} & \textcolor{cyan}{0.8138} & \textcolor{cyan}{0.9899} & \textcolor{cyan}{0.8579} & \textcolor{red}{0.9899} & \textcolor{cyan}{0.9766} & \textcolor{cyan}{0.8507} & \textcolor{cyan}{0.9875}\\
\textbf{CGDD-Net} & \textcolor{red}{\textbf{0.8421}} & \textbf{0.9768} & \textcolor{red}{\textbf{0.9704}} & \textcolor{red}{\textbf{0.8323}} & \textcolor{red}{\textbf{0.9824}} & \textcolor{red}{\textbf{0.8600}} & \textbf{0.9815} & \textcolor{red}{\textbf{0.9752}} & \textbf{0.8102} & \textcolor{red}{\textbf{0.9938}} & \textcolor{red}{\textbf{0.8661}} & \textbf{0.9812} & \textcolor{red}{\textbf{0.9775}} & \textcolor{red}{\textbf{0.8510}} & \textcolor{red}{\textbf{0.9895}}\\
\bottomrule
\end{tabular*}
\end{table*}

Relative to the baseline, the complete network increases F1 from 0.8075 to 0.8323 on DRIVE, from 0.8041 to 0.8102 on CHASE\_DB1, and from 0.8161 to 0.8510 on STARE. The corresponding gains are 2.48, 0.61, and 3.49 percentage points. AUC improves by 0.23, 1.21, and 2.11 points, and SE by 8.84, 5.37, and 8.51 points. The full model therefore improves vessel recall, foreground overlap, and probability discrimination over the baseline on all three datasets. The intermediate configurations show how these gains develop. CSDE increases SE on all three datasets. With SAMG added, SE rises further to 0.8157, 0.8527, and 0.8431, consistent with the use of spatially selected receptive-field responses. Adding DCDF increases AUC to 0.9821, 0.9870, and 0.9859. Detail-guided decoding then raises F1 to 0.8301, 0.8197, and 0.8424. These comparisons link the complete architecture to successive changes in feature extraction, fusion, and reconstruction.

Selective skips and CCA determine the final balance between metrics. On DRIVE and STARE, the complete model has the highest F1 in the ablation table. On CHASE\_DB1, the detail-guided decoder configuration reaches the highest F1, 0.8197, whereas the complete model reaches the highest AUC, 0.9938, and SE, 0.8600. In the last step, CCA increases CHASE\_DB1 AUC from 0.9899 to 0.9938 while F1 changes from 0.8138 to 0.8102.

The cumulative sequence therefore changes the balance between probability discrimination and fixed-threshold overlap.

\begin{figure*}[pos=t]
\centering\includegraphics[width=\textwidth]{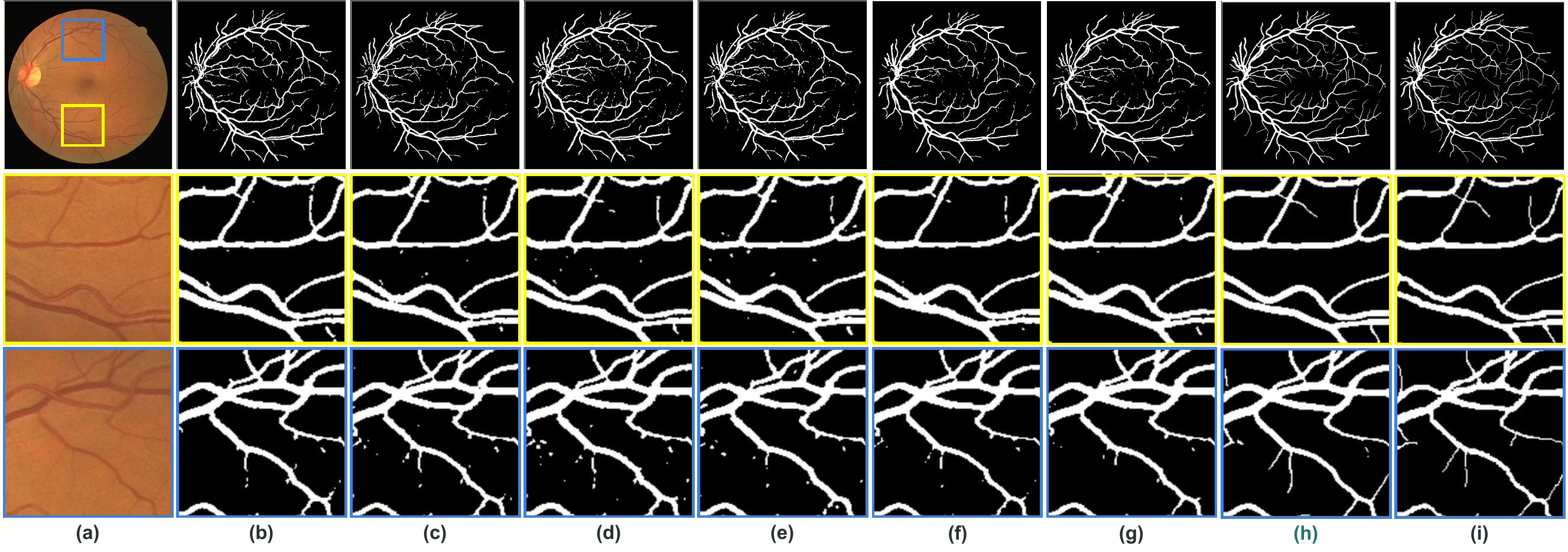}
\caption{Qualitative comparison of the cumulative configurations in Table~\ref{tab:ablation_drive}. Columns show (a) the fundus image, (b) the baseline, (c) +CSDE, (d) +SAMG, (e) +DCDF, (f) +Detail-Guided Decoder, (g) +Selective Skip, (h) the full CGDD-Net with CCA, and (i) ground truth. Configurations (c)--(h) cumulatively extend the preceding model. The top row shows the full-image results; the middle and bottom rows magnify the yellow- and blue-boxed regions, respectively.}
\label{fig:ablation}
\end{figure*}

Fig.~\ref{fig:ablation} visualizes the same sequence. The enlarged regions show changes in branch delineation and local background responses as adaptive encoding and shared detail processing are introduced. Together with Table~\ref{tab:ablation_drive}, these examples relate the architectural progression to its quantitative and spatial behavior.

\subsection{Cross-dataset evaluation}\label{sec:crossdataset}
We evaluate all twelve directed pairs among DRIVE (D), CHASE\_DB1 (C), STARE (S), and HRF (H). A source-trained model is applied directly to the target dataset, without target-domain fine-tuning, threshold tuning, or normalization adaptation. The grayscale scaling, fixed 0.5 threshold, and FOV-aware sliding-window reconstruction are retained. Predictions are reconstructed at native image size. This is a direct transfer setting, distinct from domain adaptation \cite{Galdran2022Minimalistic}.

\begin{table*}[pos=t]
\caption{Direct zero-shot cross-dataset performance of CGDD-Net. Each entry reports the target score (\%) followed by its difference from the corresponding target-domain reference (percentage points). No target-domain fine-tuning, threshold tuning, or normalization adaptation is used. D=DRIVE, C=CHASE\_DB1, S=STARE, and H=HRF.}
\label{tab:cross_validation}
\centering
\CGDDTableStyle
\setlength{\tabcolsep}{4.0pt}
\renewcommand{\arraystretch}{1.08}
\begin{tabular*}{\textwidth}{@{\extracolsep{\fill}}lccccc}
\toprule
\CGDDTableHeadStrut Dataset & F1 & ACC & AUC & SE & SP\\
\midrule
D $\rightarrow$ C & 71.44 (\textcolor{cyan}{-9.58}) & 96.11 (\textcolor{cyan}{-1.41}) & 99.21 (\textcolor{cyan}{-0.17}) & 79.64 (\textcolor{cyan}{-6.36}) & 97.11 (\textcolor{cyan}{-1.04})\\
D $\rightarrow$ S & 82.49 (\textcolor{cyan}{-2.61}) & 97.25 (\textcolor{cyan}{-0.50}) & 98.90 (\textcolor{cyan}{-0.05}) & 87.27 (\textcolor{red}{+0.66}) & 97.59 (\textcolor{cyan}{-0.53})\\
D $\rightarrow$ H & 77.61 (\textcolor{cyan}{-3.96}) & 96.11 (\textcolor{cyan}{-1.00}) & 98.90 (\textcolor{red}{+0.16}) & 87.84 (\textcolor{red}{+4.22}) & 96.80 (\textcolor{cyan}{-1.43})\\
\midrule
C $\rightarrow$ D & 68.94 (\textcolor{cyan}{-14.29}) & 95.67 (\textcolor{cyan}{-1.37}) & 80.05 (\textcolor{cyan}{-18.19}) & 55.39 (\textcolor{cyan}{-28.82}) & 99.26 (\textcolor{red}{+1.58})\\
C $\rightarrow$ S & 76.61 (\textcolor{cyan}{-8.49}) & 97.00 (\textcolor{cyan}{-0.75}) & 86.25 (\textcolor{cyan}{-12.70}) & 66.28 (\textcolor{cyan}{-20.33}) & 99.10 (\textcolor{red}{+0.98})\\
C $\rightarrow$ H & 79.06 (\textcolor{cyan}{-2.51}) & 96.90 (\textcolor{cyan}{-0.21}) & 96.11 (\textcolor{cyan}{-2.63}) & 76.24 (\textcolor{cyan}{-7.38}) & 98.65 (\textcolor{red}{+0.42})\\
\midrule
S $\rightarrow$ D & 78.35 (\textcolor{cyan}{-4.88}) & 96.59 (\textcolor{cyan}{-0.45}) & 90.67 (\textcolor{cyan}{-7.57}) & 71.15 (\textcolor{cyan}{-13.06}) & 98.70 (\textcolor{red}{+1.02})\\
S $\rightarrow$ C & 72.08 (\textcolor{cyan}{-8.94}) & 96.51 (\textcolor{cyan}{-1.01}) & 98.37 (\textcolor{cyan}{-1.01}) & 74.02 (\textcolor{cyan}{-11.98}) & 97.94 (\textcolor{cyan}{-0.21})\\
S $\rightarrow$ H & 78.58 (\textcolor{cyan}{-2.99}) & 96.51 (\textcolor{cyan}{-0.60}) & 99.01 (\textcolor{red}{+0.27}) & 83.58 (\textcolor{cyan}{-0.04}) & 97.60 (\textcolor{cyan}{-0.63})\\
\midrule
H $\rightarrow$ D & 79.96 (\textcolor{cyan}{-3.27}) & 96.74 (\textcolor{cyan}{-0.30}) & 93.01 (\textcolor{cyan}{-5.23}) & 74.69 (\textcolor{cyan}{-9.52}) & 98.36 (\textcolor{red}{+0.68})\\
H $\rightarrow$ C & 74.28 (\textcolor{cyan}{-6.74}) & 96.83 (\textcolor{cyan}{-0.69}) & 97.99 (\textcolor{cyan}{-1.39}) & 74.98 (\textcolor{cyan}{-11.02}) & 98.19 (\textcolor{red}{+0.04})\\
H $\rightarrow$ S & 83.25 (\textcolor{cyan}{-1.85}) & 97.59 (\textcolor{cyan}{-0.16}) & 96.17 (\textcolor{cyan}{-2.78}) & 80.75 (\textcolor{cyan}{-5.86}) & 98.36 (\textcolor{red}{+0.24})\\
\bottomrule
\end{tabular*}
\end{table*}

Each parenthesized value is the difference between the transfer score and the corresponding target-domain CGDD-Net reference, expressed in percentage points. The reference model uses target-domain supervision, while the transfer model uses source-domain training. The table therefore describes the effect of changing the training domain within CGDD-Net.

The strongest transferred F1 is 83.25\% for H$\rightarrow$S, followed by 82.49\% for D$\rightarrow$S.

Several transfers retain high AUC: D$\rightarrow$C reaches 99.21\%, S$\rightarrow$H reaches 99.01\%, and D$\rightarrow$S and D$\rightarrow$H both reach 98.90\%. The D-trained and S-trained HRF transfers exceed the target-domain AUC reference by 0.16 and 0.27 percentage points, respectively.

Transfer behavior depends on direction. D$\rightarrow$C retains AUC 99.21\%, whereas C$\rightarrow$D gives AUC 80.05\%, SE 55.39\%, and SP 99.26\%. F1 is lower than the target-domain reference in all twelve directions, with decreases ranging from 1.85 to 14.29 percentage points. The complete transfer matrix distinguishes retained probability discrimination from changes in the binary vessel mask under a fixed operating threshold.

\section{Discussion}\label{sec:discussion}

\begin{figure*}[pos=t]
\centering\includegraphics[width=.86\textwidth]{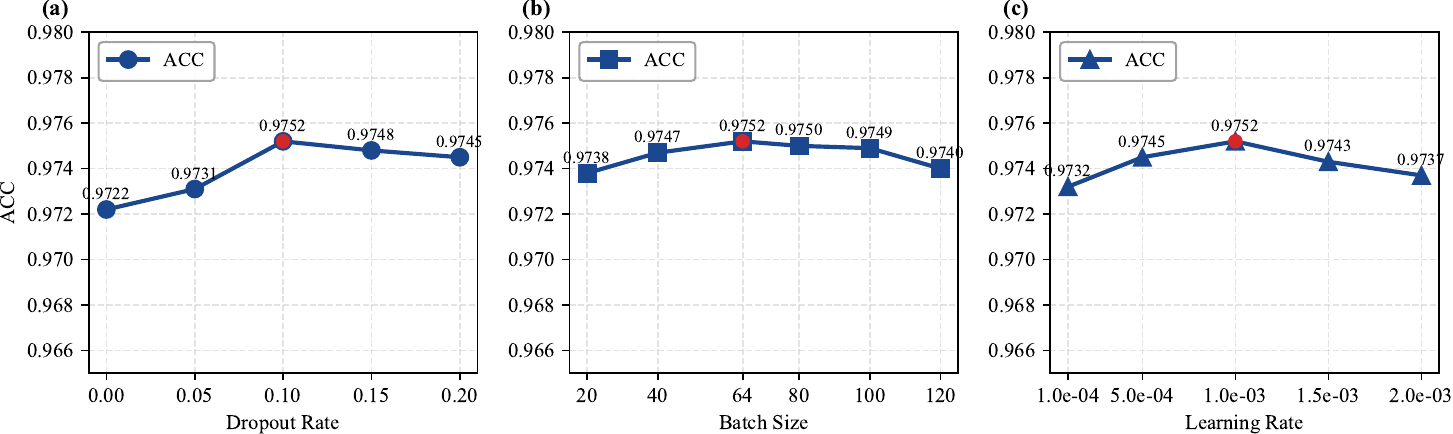}
\caption{Hyperparameter sensitivity on CHASE\_DB1. Accuracy (ACC) is plotted against (a) dropout rate, (b) batch size, and (c) learning rate, varying one factor while keeping the remaining settings fixed. Red markers identify the configuration used in the main experiments: dropout $0.1$, batch size $64$, and learning rate $10^{-3}$. Labels above the markers give the corresponding ACC values.}
\label{fig:stability}
\end{figure*}

\subsection{Contribution of shared detail-guided reconstruction}
The central design of CGDD-Net is the organization of intermediate encoder information before it enters the decoder. CSDE adapts the spatial support of the encoder features, SAMG selects among receptive-field responses, and DCDF aggregates three intermediate levels. The decoder reuses this fused representation through independent stage-specific projections, separating cross-scale consolidation from resolution-specific reconstruction. The selective skips complement the shared pathway. The direct $E_1$ connection supplies full-resolution localization, while $E_4$ contributes deeper context. Features from $E_2$ and $E_3$ enter reconstruction through the shared representation. Consequently, the decoder receives selected multi-scale information without carrying each intermediate feature map through a separate direct skip.

The ablation results connect this organization to the observed improvements. The complete network increases SE, F1, and AUC over the internal baseline on all three evaluated datasets. The intermediate rows show gains as adaptive encoding is connected to gated fusion and detail-guided decoding. The different F1 and AUC optima on CHASE\_DB1 also clarify the role of the final refinement: improved probability discrimination need not coincide with the highest overlap at one threshold.

These experiments assess the cumulative architecture. Their gains describe the complete feature-extraction and reconstruction pathway.

\subsection{Hyperparameter response}
Fig.~\ref{fig:stability} examines one factor at a time on CHASE\_DB1. Increasing dropout from 0 to 0.1 raises ACC from 0.9722 to 0.9752, with little further change up to 0.2. Batch sizes between 20 and 120 yield ACC values from 0.9738 to 0.9752. Over learning rates from $10^{-4}$ to $2\times10^{-3}$, the ACC range is 0.0020 and its peak occurs at $10^{-3}$.

The default configuration lies within these tested ranges: dropout 0.1, batch size 64, and learning rate 0.001. The small ACC changes describe local one-factor responses under the stated schedule and complement the architectural comparisons in Table~\ref{tab:ablation_drive}.

\subsection{Model size and computational profile}
Table~\ref{tab:complexity} relates model size to STARE segmentation performance. CGDD-Net uses 1.96M parameters, compared with 34.53M for the listed U-Net and 2.83M for MDF-Net, corresponding to reductions of approximately 94.3\% and 30.7\%. Among the displayed methods, CGDD-Net has the second-smallest parameter count and the second-highest F1 and AUC. Its compactness comes from the narrow encoder widths and the fixed-width intermediate representation reused during decoding.

\begin{table}[pos=t]
\caption{Model complexity and segmentation performance on STARE. AUC and F1 are percentages. Parameter counts and FLOPs are reported in millions and billions, respectively. Red and cyan identify the best and second-best values in each column. FLOPs are retained as reported model-forward counts, not as measurements of full-image latency under a common profiling configuration.}
\label{tab:complexity}
\centering\CGDDTableStyle
\setlength{\tabcolsep}{4.2pt}
\begin{tabular*}{\columnwidth}{@{\extracolsep{\fill}}lrrrr}
\toprule
Method & AUC (\%) & F1 (\%) & Parameters (M) & FLOPs (G)\\
\midrule
U-Net \cite{Ronneberger2015UNet} & 97.37 & 76.55 & 34.53 & 50.11\\
AttU-Net \cite{Oktay2018AttentionUNet} & 97.75 & 75.43 & 34.88 & 50.96\\
IterNet \cite{Li2020IterNet} & 98.81 & 81.46 & \textcolor{red}{1.36} & \textcolor{red}{19.44}\\
R2U-Net \cite{Alom2018R2UNet} & 97.61 & 77.48 & 39.09 & 117.08\\
AMFEF-Net \cite{Zhu2025AMFEF} & 98.39 & 79.85 & 36.79 & 52.33\\
MDF-Net \cite{Ma2025MDFNet} & \textcolor{red}{99.36} & \textcolor{red}{87.03} & 2.83 & 46.36\\
\midrule
\textbf{CGDD-Net (Ours)} & \textbf{\textcolor{cyan}{98.95}} & \textbf{\textcolor{cyan}{85.10}} & \textbf{\textcolor{cyan}{1.96}} & \textbf{\textcolor{cyan}{29.96}}\\
\bottomrule
\end{tabular*}
\end{table}

IterNet uses fewer parameters but has lower reported STARE F1 and AUC; MDF-Net achieves higher scores with more parameters. CGDD-Net lies between them in both model size and segmentation performance.

The listed 29.96 GFLOPs describes reported model-forward computation. Full-image processing additionally depends on image dimensions, window overlap, memory transfer, and stitching. For the reconstruction procedure in Section~\ref{sec:implementation}, a dimension of length $L$ uses $1+\max(0,\lceil(L-96)/16\rceil)$ windows. The architecture-level counts and the reconstruction procedure therefore describe different parts of the computational workload; the table is not interpreted as a hardware timing comparison.

\begin{figure*}[t]
\centering
\includegraphics[width=.96\textwidth]{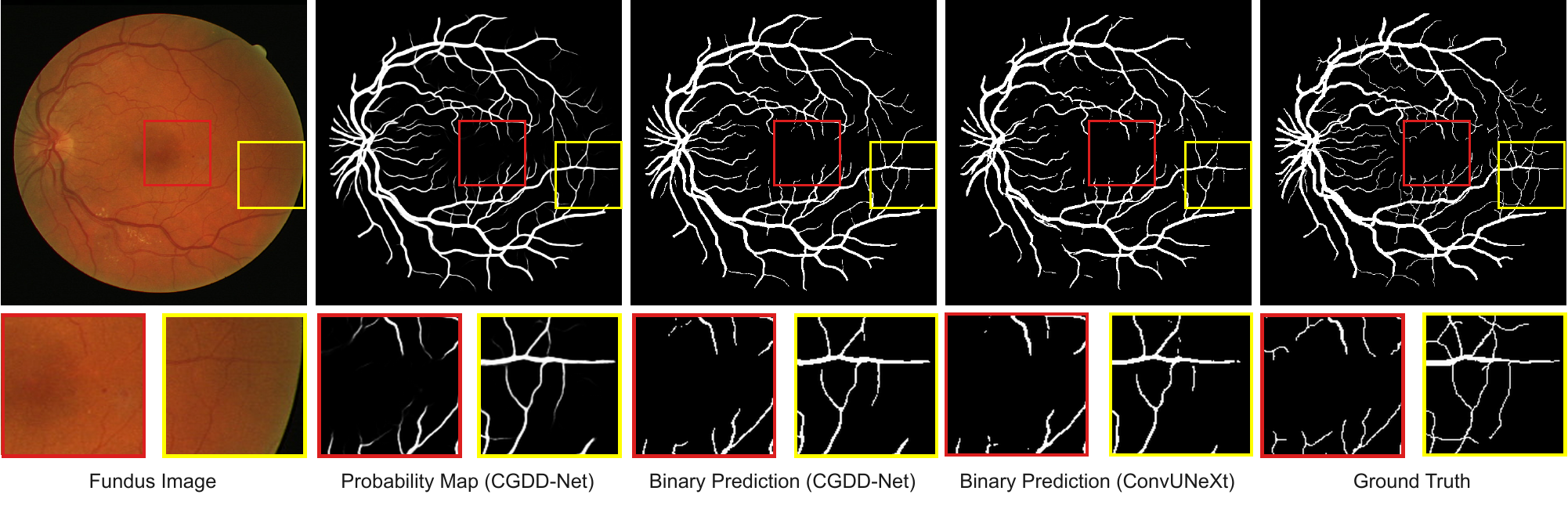}
\caption{Probability and binary outputs for a low-contrast retinal image. From left to right: the fundus image, the CGDD-Net probability map, its binary segmentation at threshold $0.5$, the ConvUNeXt binary segmentation, and ground truth. Brighter values in the probability map indicate stronger vessel responses. Paired crops below each image enlarge the red- and yellow-boxed regions, allowing faint responses to be inspected before and after thresholding.}
\label{fig:limitations}
\end{figure*}

\subsection{Cross-domain behavior and probability reconstruction}
The cross-dataset experiments characterize how the learned representation behaves when the source domain changes. High AUC is retained in several directions, especially D$\rightarrow$C and the D- and S-trained transfers to HRF. The asymmetry between D$\rightarrow$C and C$\rightarrow$D shows that transfer is specific to the source-target pair. Differences in image appearance, resolution, and annotation conventions provide relevant context for this behavior \cite{Lyu2022RETA}.

AUC and F1 describe different output properties. AUC measures how probabilities rank vessel and background pixels across thresholds, whereas F1 evaluates the mask at 0.5. Consequently, fewer vessel pixels may exceed that fixed threshold after transfer even when the ranking score remains high.

Fig.~\ref{fig:limitations} illustrates the two output stages on a low-contrast image. The probability map preserves graded responses along faint branches, while thresholding retains only responses at or above the operating point. The red and yellow regions place these responses beside the binary predictions and the reference annotation. This view complements the transfer matrix by showing how image-level probability reconstruction relates to local vessel delineation.

\section{Conclusion}\label{sec:conclusion}
We proposed CGDD-Net for retinal vessel segmentation through context-guided extraction and shared detail-guided reconstruction. Deformable local encoding and spatial multi-kernel selection feed a common eight-channel representation, which combines three intermediate encoder levels and is reused at three decoder resolutions.

Experiments on DRIVE, CHASE\_DB1, STARE, and HRF demonstrate competitive segmentation with 1.96M trainable parameters. Cumulative ablations show gains in vessel recall, F1, and AUC over the internal baseline on all three tested datasets, while twelve directed transfers characterize behavior beyond the training domain. The work establishes a compact way to connect adaptive intermediate features to successive reconstruction stages.

\section*{CRediT authorship contribution statement}
\textbf{Xincheng Li:} Conceptualization, Methodology, Software, Validation, Formal analysis, Investigation, Data curation, Visualization, Writing -- original draft, Writing -- review \& editing. \textbf{Xiaoqi Sheng:} Supervision, Methodology, Validation, Writing -- review \& editing. \textbf{Xinyu Zhang:} Validation, Formal analysis, Investigation, Writing -- review \& editing.

\section*{Declaration of competing interest}
The authors declare that they have no known competing financial interests or personal relationships that could have appeared to influence the work reported in this paper.

\section*{Funding}
This research did not receive any specific grant from funding agencies in the public, commercial, or not-for-profit sectors.

\section*{Ethics statement}
This study uses public retrospective retinal image datasets and involves no new participant recruitment or intervention. Image and annotation reuse is subject to the terms of the respective dataset providers.

\section*{Data and code availability}
The fundus images and reference annotations are available from the original providers of DRIVE~\cite{Staal2004DRIVE}, CHASE\_DB1~\cite{Fraz2012CHASE}, STARE~\cite{Hoover2000STARE}, and HRF~\cite{Budai2013HRF}, subject to their access and reuse terms. A reference implementation of CGDD-Net is available at \url{https://github.com/lixincheng-xcl/CGDD-Net}; the source snapshot referenced here is commit \texttt{6d853394f391}.

\FloatBarrier
\bibliographystyle{elsarticle-num}
\bibliography{references}
\end{document}